\pdfoutput=1
\documentclass[11pt,a4paper]{article}
\usepackage[hyperref]{acl2021}
\usepackage{times}
\usepackage{latexsym}
\usepackage{amsmath}
\usepackage{graphicx}
\usepackage{xcolor}
\usepackage{subfigure}
\usepackage{booktabs}
\usepackage{multirow}
\usepackage{multicol}

\usepackage[T1]{fontenc}
\usepackage[utf8]{inputenc}

\usepackage{microtype}

\title{Models In a Spelling Bee:\\
Language Models Implicitly Learn the Character Composition of Tokens}

\author{Itay Itzhak \qquad Omer Levy \\
The Blavatnik School of Computer Science \\
Tel Aviv University \\
\texttt{\{itay1itzhak,omerlevy\}@gmail.com}}

\date{}

\begin{document}
\maketitle

\begin{abstract}
Standard pretrained language models operate on sequences of subword tokens without direct access to the characters that compose each token's string representation.
We probe the embedding layer of pretrained language models and show that models learn the internal character composition of whole word and subword tokens to a surprising extent, without ever seeing the characters coupled with the tokens.
Our results show that the embedding layers of RoBERTa and GPT2 each hold enough information to accurately spell up to a third of the vocabulary and reach high character $n$gram overlap across all token types.
We further test whether enriching subword models with character information can improve language modeling, and observe that this method has a near-identical learning curve as training without spelling-based enrichment.
Overall, our results suggest that language modeling objectives incentivize the model to implicitly learn some notion of spelling, and that explicitly teaching the model how to spell does not appear to enhance its performance on such tasks.\footnote{Our code is available at: \url{https://github.com/itay1itzhak/SpellingBee}}
\end{abstract}
\section{Introduction}

Contemporary subword tokenization algorithms such as BPE \citep{sennrich-etal-2016-neural} partition a string into contiguous spans of characters.
Each span represents a frequent character $n$gram, from individual characters (\textit{a}), through prefixes (\textit{uni}) and suffixes (\textit{tion}), and even complete words (\textit{cats}).
The tokenizer then converts each such span into a discrete symbol (a token) with no internal structure, effectively discarding the token's orthographic information.
Therefore, a model operating over sequences of subword tokens should be oblivious to the spelling of each token.
In this work, we show that despite having no direct access to the subwords' internal character composition, pretrained language models \textit{do} learn some notion of spelling.

To examine what pretrained language models learn about spelling, we present the \textit{SpellingBee} probe. 
SpellingBee is a generative language model that predicts the character composition of a token given only its (uncontextualized) vector representation from the pretrained model's embeddings matrix.
SpellingBee is trained on part of the model's vocabulary, and then tested by spelling unseen token types.
If the probe can successfully reconstruct the correct character sequence from an unseen token's embedding, then there must be significant orthographic information encoded in the vector.

We find that the embedding layers of several pretrained language models contain surprising amounts of character information. 
SpellingBee accurately spells 31.8\% of the held-out vocabulary for RoBERTa-Large \citep{liu2019roberta}, 32.9\% for GPT2-Medium \citep{radford2019language}, and 40.9\% for the Arabic language model AraBERT-Large \citep{antoun-etal-2020-arabert}.
A softer metric that is sensitive to partially-correct spellings (chrF) \cite{popovic-2015-chrf} shows a similar trend, with 48.7 for RoBERTa-Large and 62.3 for AraBERT-Large.
These results are much higher than the baseline of applying SpellingBee to randomly-initialized vectors, which fails to spell a single token.

Given that subword models learn some notion of character composition to fulfill language modeling objectives, could they perhaps benefit from knowing the exact spelling of each token a priori?
To that end, we reverse SpellingBee's role and use it to pretrain the embedding layer of a randomly-initialized model, thus imbuing each token representation with its orthographic information before training the whole model on the masked language modeling objective.
We compare the pretraining process of the character-infused model to that of an identical model whose embedding layer is randomly initialized (and not pretrained), and find that both learning curves converge to virtually identical values within the first 1,000 gradient updates, a fraction of the total optimization process.
This experiment suggests that while language models may need to learn some notion of spelling to optimize their objectives,
they might also be able to quickly acquire most of the character-level information they need from plain token sequences without directly observing the composition of each token.

\section{Spelling Bee}
\label{sec:model}

To measure how much a model knows the character composition of its tokens,
we introduce SpellingBee, a generative probe that tries to spell out a token character-by-character.
Specifically, SpellingBee probes the original model's \textit{embedding matrix}, since spelling is a property of token \textit{types}, invariant to context.
For example, given the embedding of the token \textit{cats}, SpellingBee will try to generate the sequence [\textit{c}, \textit{a}, \textit{t}, \textit{s}].
We do so by modeling SpellingBee as a character-based language model,\footnote{Implemented using the transformer decoder architecture, following standard practice in language modeling.} where the first token is a vector representation of the vocabulary item.\footnote{Some vocabularies have symbols for indicating preceding whitespaces (\textit{\_}) or that the next token is part of the same word (\textit{\#\#}). SpellingBee learns to predict these symbols too.}

\paragraph{Training}
We split the vocabulary to train and test sets,\footnote{We test various train/test splits to ensure the robustness of our findings. See Section~\ref{sec:experiments} for more detail.} 
and use teacher forcing to train SpellingBee.
In the example of \textit{cats}, SpellingBee will compute the following probabilities: 
\begin{align*}
P(x_1 = \text{\textit{c}} & \mid x_0 = \text{\textit{cats}}) \\
P(x_2 = \text{\textit{a}} & \mid x_0 = \text{\textit{cats}}, x_1 = \text{\textit{c}}) \\
P(x_3 = \text{\textit{t}} & \mid x_0 = \text{\textit{cats}}, x_1 = \text{\textit{c}}, x_2 = \text{\textit{a}}) \\
&\vdots
\end{align*}
All of SpellingBee's parameters are randomly initialized.
The only parameters that are pretrained are the token embeddings (e.g. the representation of \textit{cats} or \textit{a}), which are taken from the original pretrained language model we intend to probe, and treated as constants; i.e. kept frozen during SpellingBee's training.

\paragraph{Inference \& Evaluation}
Once SpellingBee is trained, we apply it to the test set using greedy decoding.
For each vocabulary item $w$ in the test set, SpellingBee is given only the corresponding embedding vector $e_w$, and is expected to generate the character sequence $w_1, \ldots, w_n$ that defines $w$.
We measure success on the test set using two metrics: \textit{exact match} (EM), and character $n$gram overlap score using \textit{chrF} \cite{popovic-2015-chrf}.
While EM is strict, chrF allows us to measure partial success.
We also report edit distance using Levenshtein distance ratio in Appendix \ref{sec:levenshtein}.

\paragraph{SpellingBee for Pretraining Embeddings}
While we mainly use SpellingBee as a probe, a variation of our method could potentially imbue the embedding layer with character information before training a language model.
We could train a probe with randomly-initialized embeddings (instead of pretrained embeddings from another model) to predict the spelling of \textit{all} vocabulary items,
and use these trained probe embeddings to initialize any target model's embedding layer (instead of random initialization).
We experiment with this method in Section \ref{sec:pretraining}, but find that it does not have any significant impact on the convergence of language models.

\section{Experiment Setup}
\label{sec:experiments}

We begin with a series of probing experiments, where we apply SpellingBee to the embedding layer of various pretrained models.\footnote{Hyperparameters are detailed in Appendix~\ref{sec:hyperparameters}.}

\paragraph{Pretrained Models}
We probe four pretrained models: RoBERTa-Base and Large \cite{liu2019roberta}, GPT2-Medium \cite{radford2019language}, and AraBERT-Large \cite{antoun-etal-2020-arabert}.
This set introduces some diversity in vocabulary, objective, and scale:
the first three models are trained on English corpora, while AraBERT is trained on text in Arabic;
GPT2 is an autoregressive language model, while the rest are masked language models;
RoBERTa-Base consists of 125M parameters (with 768 dimensions per embedding), while the other models have approximately 350M parameters (with 1024 dimensions per embedding).

\paragraph{Control}
Since SpellingBee is a \textit{trained} probe, we wish to establish the probe’s baseline performance when provided with inputs with no orthographic information.
As an empirical \textit{control}, we train and test SpellingBee on randomly-initialized vectors, in addition to the main experiments where we utilize the pretrained embedding layers.

\paragraph{Training \& Testing Data}
We split the vocabulary into training and testing data using the following protocol.
First, we randomly sample 1000 token types as test.
We then filter the remaining vocabulary to eliminate tokens that may be too similar to the test tokens, and randomly sample 32000 training examples.
We experiment with three filters:
\textit{none}, which do not remove tokens beyond the test-set tokens;
\textit{similarity}, which removes the top 20 most similar tokens for every token in test, according to the cosine similarity induced by the embedding vectors;
\textit{lemma}, which removes any token type that shares a lemma with a test-set token (e.g. if \textit{diving} is in the test set, then \textit{diver} cannot be in the training set).\footnote{We lemmatize using NLTK's WordNet lemmatizer \cite{loper-bird-2002-nltk} for English and Farasa's Stemmer \cite{darwish-mubarak-2016-farasa} for Arabic.}
The lemma filter always applies the similarity filter first, providing an even more adversarial approach for splitting the data.
To control for variance, we create 10 such splits for each model and filter, and report the averaged evaluation metrics over all 10 test sets.

\begin{table}[t]
\centering
\small
\begin{tabular}{@{}llcccc@{}}
\toprule
 & \multirow{2}{*}{\textbf{Filter}} & \multicolumn{2}{c}{\textbf{RoBERTa}} & \textbf{GPT2} & \textbf{AraBERT} \\
& & \textbf{Base} & \textbf{Large} & \textbf{Medium} & \textbf{Large} \\
\midrule
\multirow{4}{*}{\rotatebox[origin=c]{90}{\textbf{EM}}} & None & 27.3 & 31.8 & 32.9 & 40.9 \\
& Similarity & 15.7 & 18.2 & 17.9 & 21.9 \\
& Lemma & 15.7 & 17.7 & 16.5 & 19.5  \\
& Control & ~~0.0 & ~~0.0 & ~~0.0 & ~~0.0  \\
\midrule
\multirow{4}{*}{\rotatebox[origin=c]{90}{\textbf{chrF}}} & None & 44.7 & 48.7 & 51.6 & 62.3 \\
& Similarity & 32.7 & 35.1 & 36.4 & 46.0 \\
& Lemma & 32.6 & 34.8 & 35.2 & 43.9  \\
& Control & ~~7.0 & ~~7.0 & ~~7.0 & ~~7.0  \\
\bottomrule
\end{tabular}
\caption{The percent of token types that can be spelled out exactly (EM) from their embeddings by SpellingBee, and the $n$gram overlap between SpellingBee's reproductions and the token types' true spellings (chrF). The first three rows reflect different methods for filtering the training data, and the fourth represents the control experiment, which uses randomly initialized embeddings. All SpellingBee instances in this table are trained on 32,000 examples.}
\label{tab:results}
\end{table}

\section{Results}
\label{sec:results}

\paragraph{Main Result}
Table~\ref{tab:results} shows how well SpellingBee can spell a vocabulary token using only its frozen pretrained embedding.
We observe that SpellingBee is able to accurately recover the spelling of up to 40.9\% of the test set, while the control is unable to spell even a single word correctly. 
A similar trend can be seen when considering the finer character $n$gram metric (chrF).
Manually analyzing the predictions of the control baselines (see Appendix \ref{sec:manual_analysis}) indicate that it primarily generates combinations of frequent character sequences, which mildly contributes to the chrF score, but does not affect EM.
These results are persistent across different models and filters, strongly indicating that the embedding layer of pretrained models contains significant amounts of information about each token's character composition.

One may suggest that training SpellingBee over 32,000 examples may leak information from the test set;
for example, if \textit{dog} was seen during training, then spelling out \textit{dogs} might be easy.
We thus consider the similarity and lemma filters, which remove such near-neighbors from the training set.
While results are indeed lower (and probably do account for some level of information leakage), they are still considerably higher than the control, both in terms of EM and chrF. 
Results using the similarity and lemma filters are rather similar, suggesting that embedding-space similarity captures some information about each token's lemma.

Finally, we find that the properties of pretrained models also seem to have a significant effect on the amount of spelling information SpellingBee can extract.
Larger models tend to score higher in the probe, and the model trained on text in Arabic appears to have substantially higher EM and chrF scores than those trained on English corpora.
One possibility is that Arabic's rich morphology incentivizes the model to store more information about each token's character composition; 
however, it is also possible that AraBERT's different vocabulary, which allocates shorter character sequences to each token type, might explain this difference (we discuss the link between sequence length and accuracy in Appendix~\ref{sec:token_len}).

Overall, our probing experiments show that even though subword-based language models do not have direct access to spelling, they \textit{can} and \textit{do} learn a surprising amount of information about the character composition of each vocabulary token.

\begin{table}[t]
\centering
\small
\begin{tabular}{@{}llccc@{}}
\toprule
 & \multirow{2}{*}{\textbf{Filter}} & \multicolumn{1}{c}{\textbf{RoBERTa}} & \textbf{CharacterBERT} & \textbf{GloVe}  \\
& & \textbf{Base} & \textbf{Base} & \textbf{300D} \\
\midrule
\multirow{4}{*}{\rotatebox[origin=c]{90}{\textbf{EM}}} & None & 43.0 & 28.2 & 2.0  \\
& Similarity & 9.6 & 12.9 & 1.6  \\
& Lemma & 9.9 & 12.9 & 1.6  \\
& Control & 0.0 & 0.0 & 0.0  \\
\midrule
\multirow{4}{*}{\rotatebox[origin=c]{90}{\textbf{chrF}}} & None & 58.8 & 53.3 & 13.6  \\
& Similarity & 27.0 & 37.5 & 13.2  \\
& Lemma & 27.3 & 37.5 & 13.0   \\
& Control & 7.9 & 8.0 & 8.0  \\
\bottomrule
\end{tabular}
\caption{The percent of \textit{whole words} that can be spelled out exactly (EM) from their embeddings by SpellingBee, and the $n$gram overlap between SpellingBee's reproductions and the token types' true spellings (chrF). All SpellingBee instances in this table are trained on 32,000 examples of whole words.}
\label{tab:charcterbert_results}
\end{table}

\paragraph{Character-Aware Models}
Some models are provided with the raw character sequence of each token.
To test whether the embedding layers of such models are indeed more informed about each token's spelling,
we apply SpellingBee to CharacterBERT \cite{el-boukkouri-etal-2020-characterbert}, a BERT-style model whose layer-zero word embeddings are derived from a character CNN, following ELMo \cite{peters-etal-2018-deep}.

Table \ref{tab:charcterbert_results} shows that the spelling-aware embeddings of CharacterBERT score higher on the SpellingBee probe when the similarity and lemma filters are applied.
However, when no filter is applied, RoBERTa's character-oblivious but highly-tuned training process produces embeddings that score higher on SpellingBee, presumably by leveraging implicit similarity functions in the embedding space. 

Although CharacterBERT's embedding layer is better at reconstructing original words (when similarity filters are applied), this does not mean that character-aware models are necessarily better downstream.
\citet{el-boukkouri-etal-2020-characterbert} report performance increases only on the medical domain.
In Section~\ref{sec:pretraining}, we demonstrate that initializing a masked language model's embedding layer with character information has a negligible effect on its perplexity.

\paragraph{Context-Oblivious Models}
The first generation of neural word representations \cite{mikolov2013efficient, mikolov2013distributed} contained only embedding layers, without any contextualization mechanism.
We thus use GloVe \cite{pennington-etal-2014-glove} to estimate a lower bound on character information that can be obtained by simple context-oblivious models.
We probe the first 50K words in GloVe's vocabulary with SpellingBee.
Table \ref{tab:charcterbert_results} shows that GloVe embeddings do contain a weak orthographic signal, better than random embeddings, but substantially weaker than the information stored in the embedding layer of large transformer-based language models.



\begin{figure}[t]
\centering
\includegraphics[width=0.8\columnwidth]{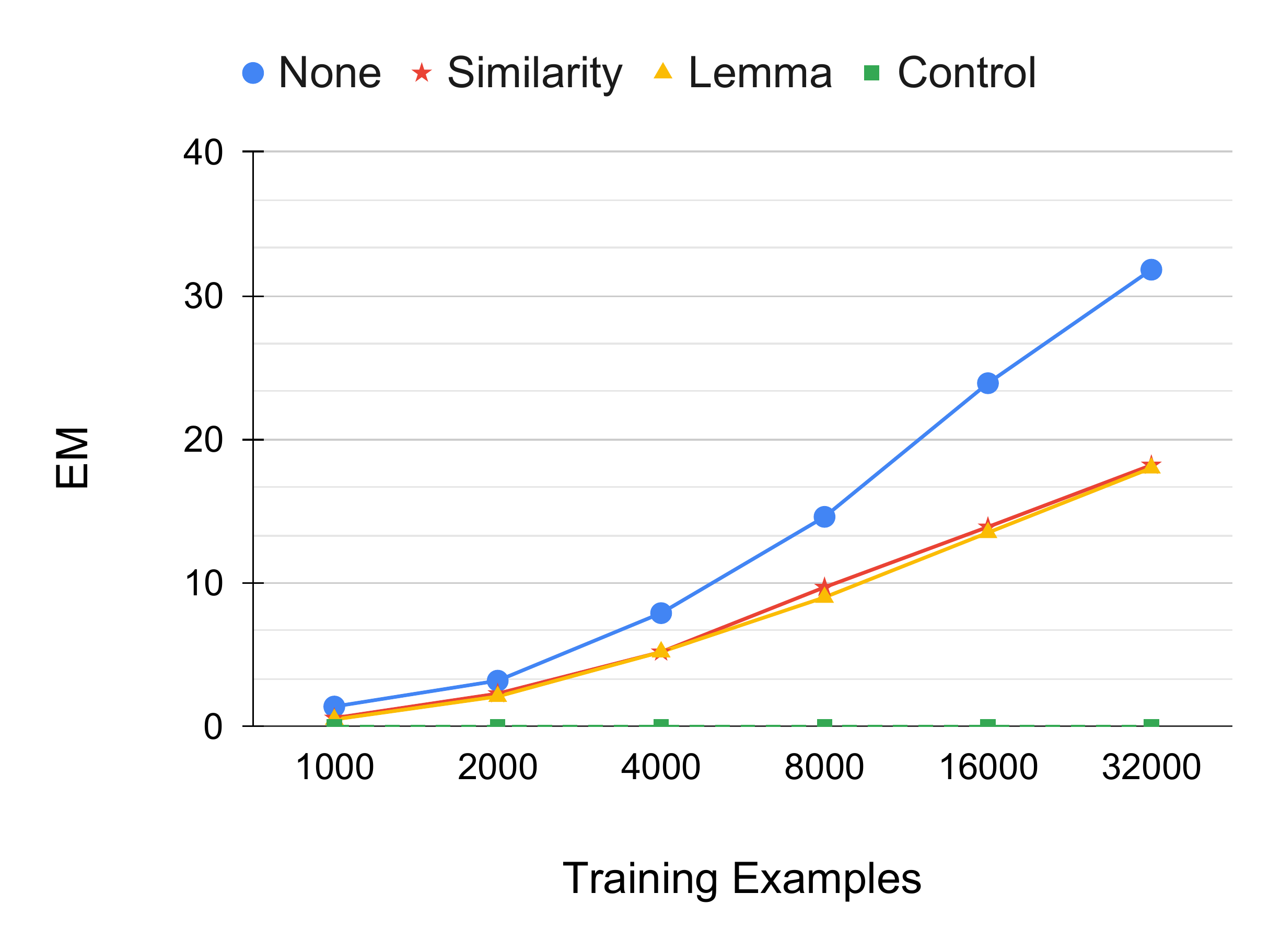}
\includegraphics[width=0.8\columnwidth]{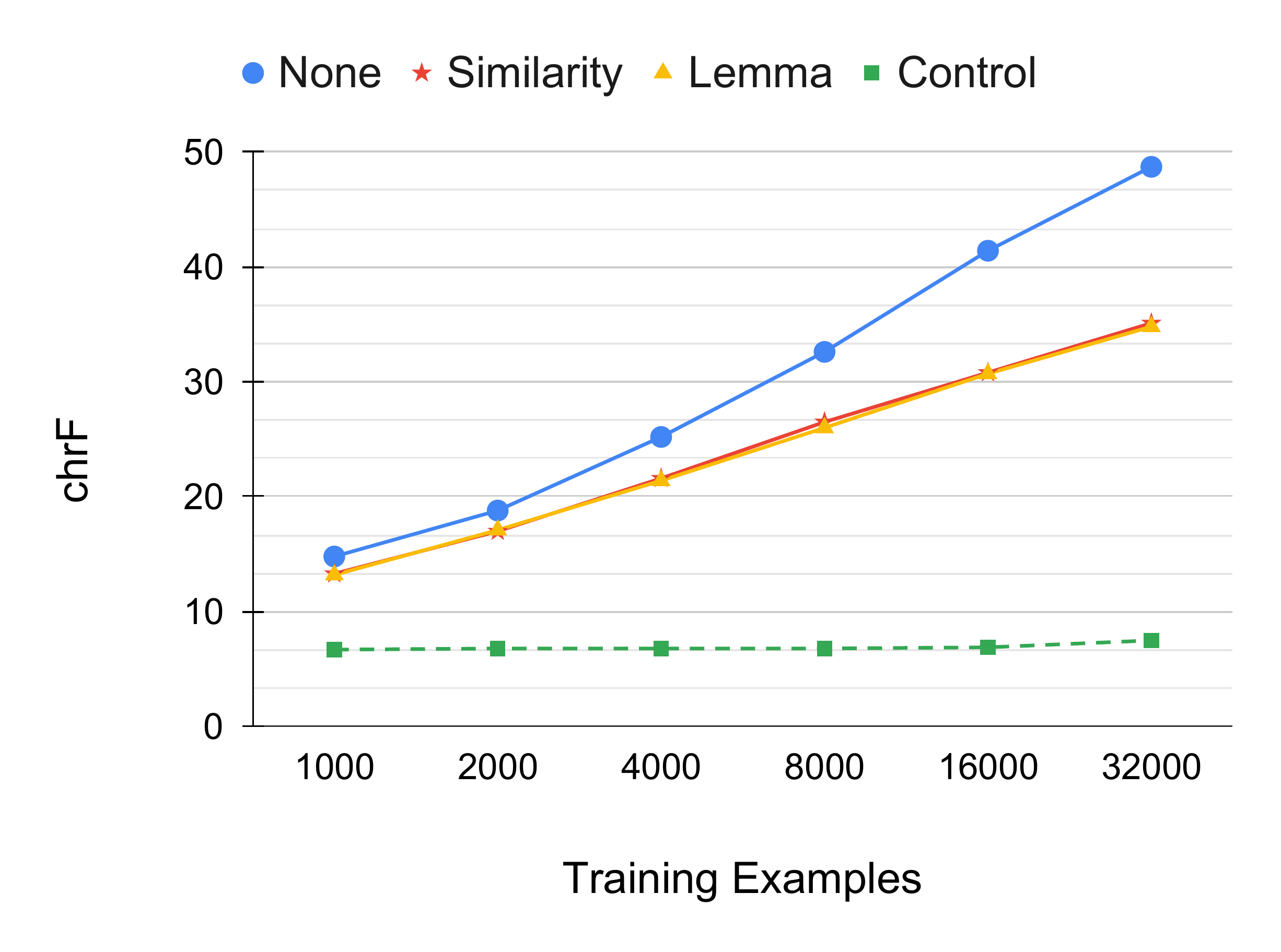}
\caption{The amount of character information SpellingBee is able to extract from RoBERTa-Large, as measured by EM (top) and chrF (bottom), given different quantities of training examples.}
\label{fig:examples_curve}
\end{figure}

\paragraph{Probing with Less Training Data}
We further examine whether SpellingBee can extract information when trained on less examples.
Figure~\ref{fig:examples_curve} shows how well SpellingBee can spell RoBERTa-Large's vocabulary when trained on varying amounts of data, across all filters.
We find that more data makes for a better probe, but that even a few thousand examples are enough to train SpellingBee to extract significant character information from the embeddings, which \textit{cannot} be extracted from randomized vectors (the control).\footnote{We provide additional analysis on spelling accuracy by subword frequency and length in Appendices~\ref{sec:token_freq} and \ref{sec:token_len}.}

\begin{figure*}[th!]
\centering
\includegraphics[width=0.68\columnwidth]{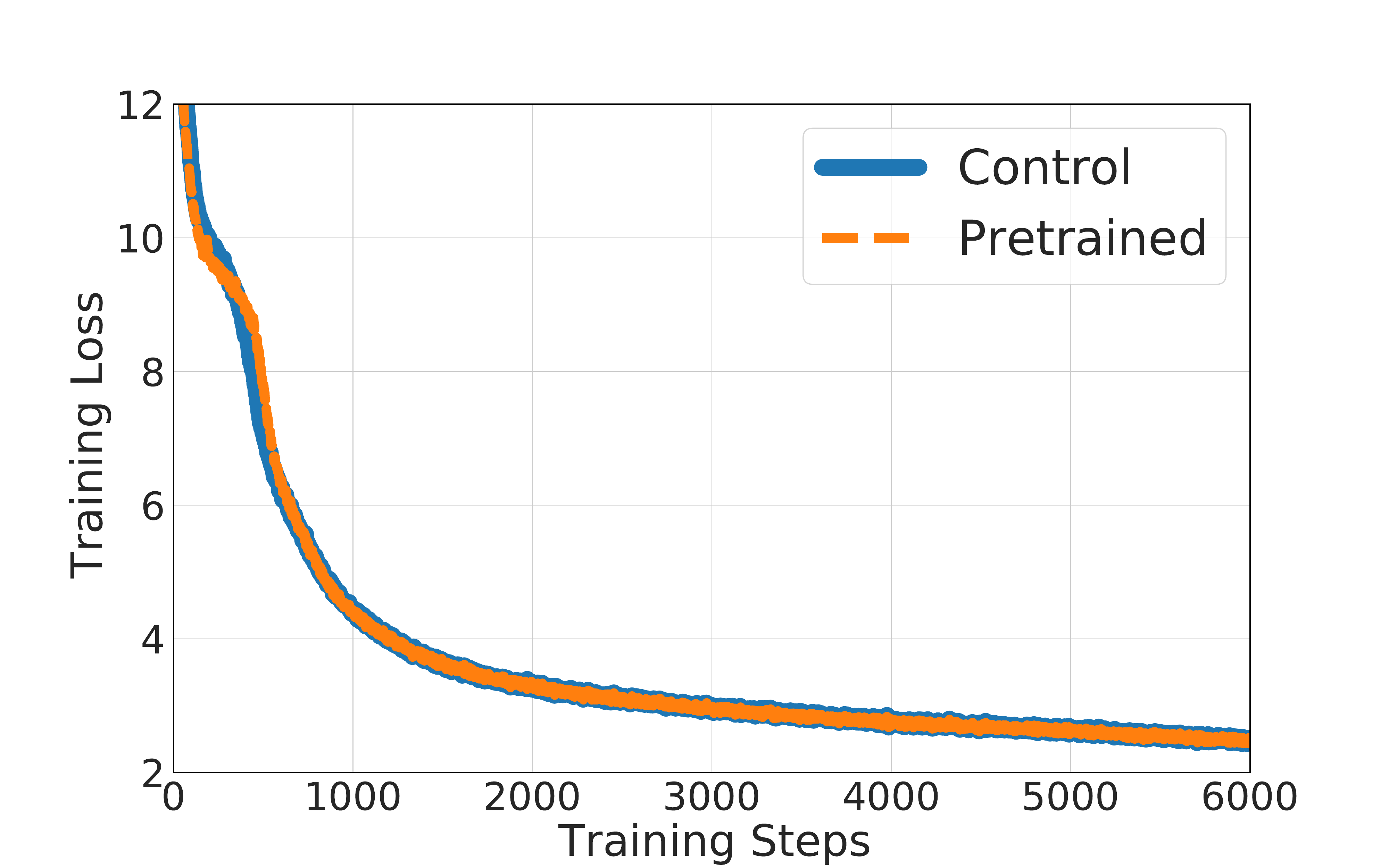}
\hfill
\includegraphics[width=0.68\columnwidth]{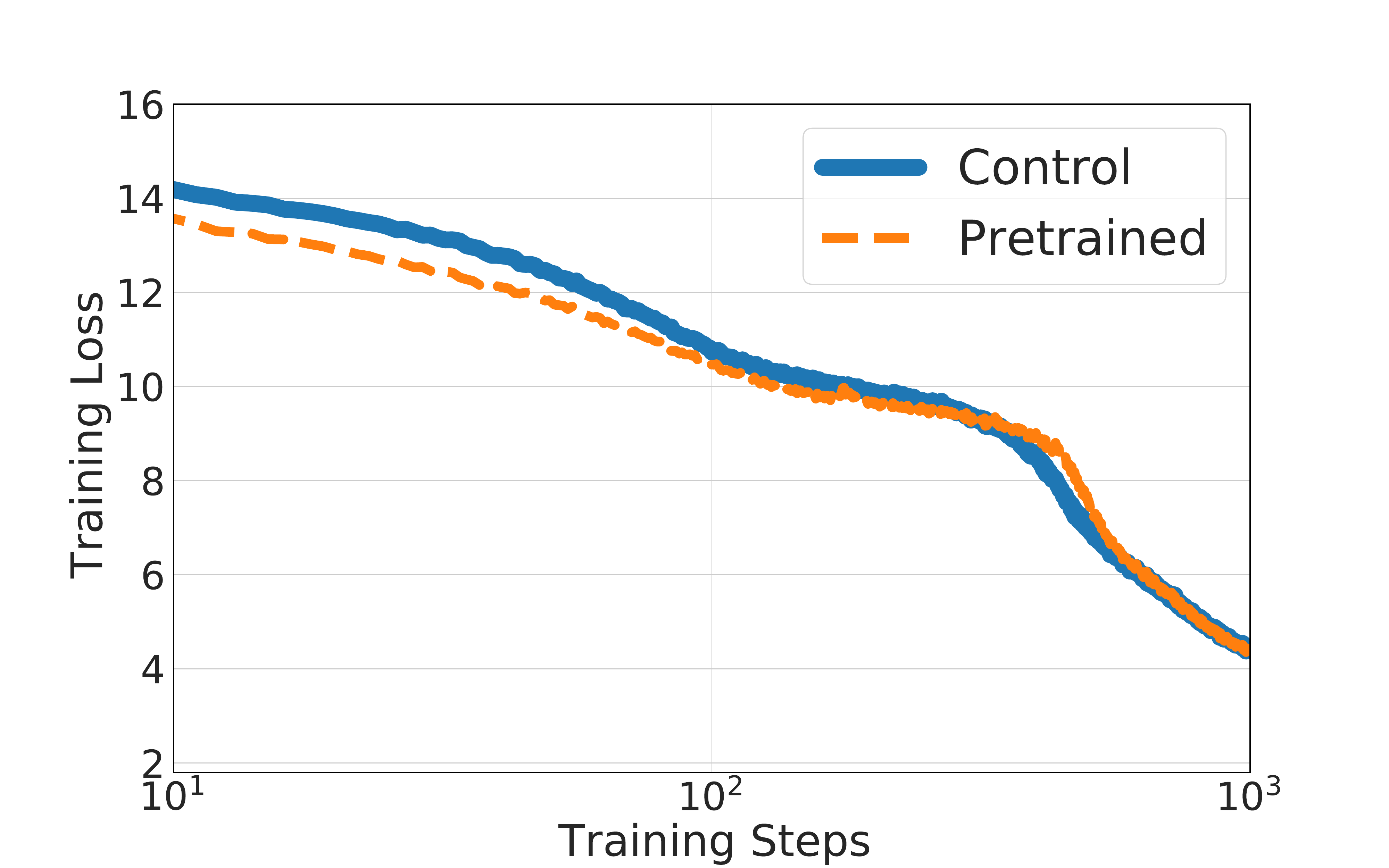}
\hfill
\includegraphics[width=0.68\columnwidth]{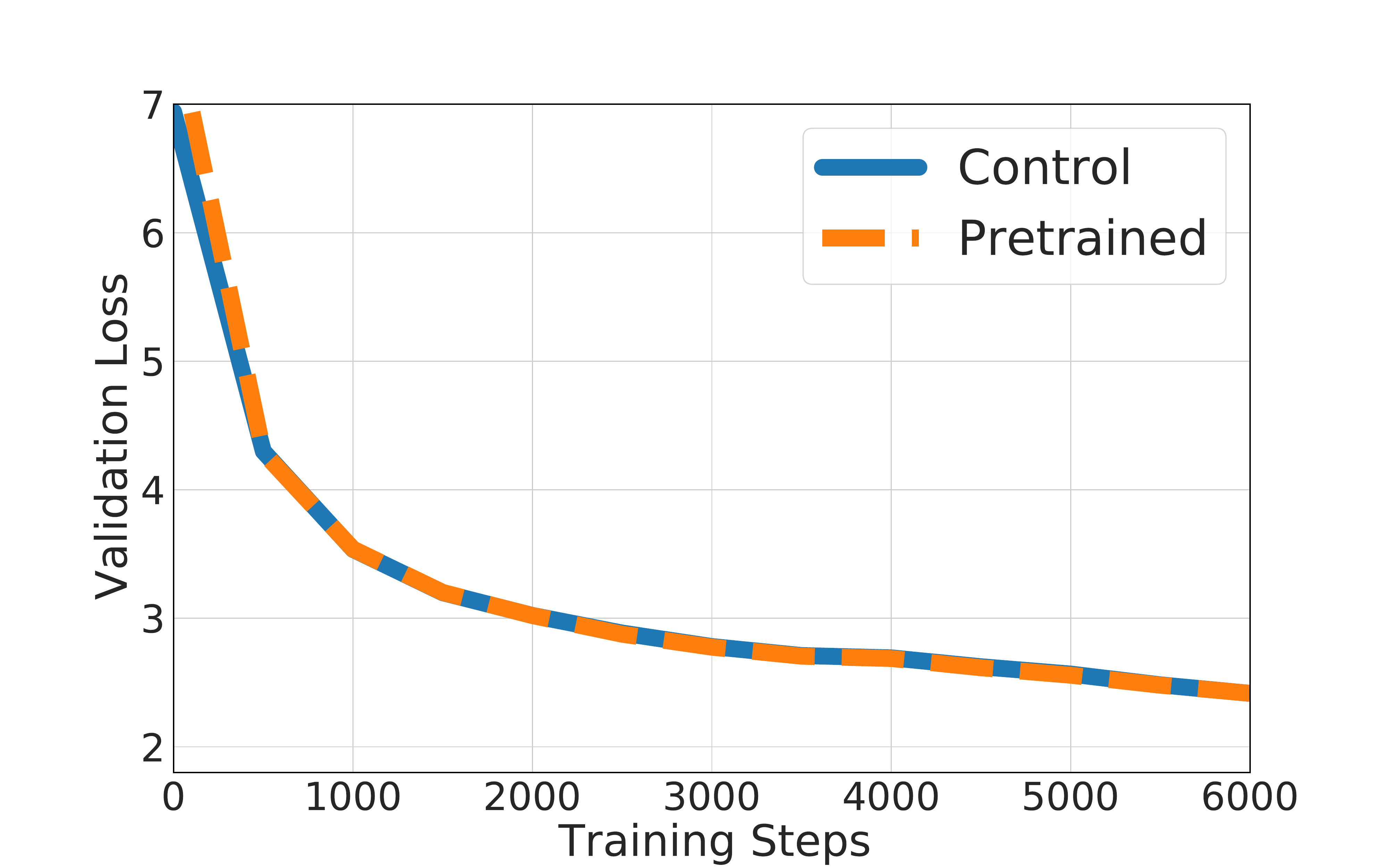}
\caption{The overall training loss (left), first steps of training loss (center), and validation loss (right) of RoBERTa-Large, when training on the masked language modeling objective with embeddings pretrained by SpellingBee (\textit{pretrained}) and randomly-initialized embeddings (\textit{control}).}
\label{fig:pretraining}
\end{figure*}

\section{Pretraining Language Models to Spell}
\label{sec:pretraining}

Our probing experiments reveal that language models learn some partial notion of spelling, despite the lack of direct access to characters.
Therefore, we hypothesize that learning to spell is beneficial for language models, and propose pretraining the embedding layer using a variant of the SpellingBee probe described in Section~\ref{sec:model}.
Here, the goal is to imbue each embedding with enough information for SpellingBee to accurately generate its surface form, and then initialize the language model with the pretrained embeddings before it starts training on the language modeling objective.

We apply this process to RoBERTa-Large, training the model's embedding layer with SpellingBee using the same hyperparameter settings from Appendix~\ref{sec:hyperparameters}, with the key difference being that the embeddings are now tunable parameters (not frozen).\footnote{To verify that this process does indeed encode the tokens' spellings into the embeddings, we apply a SpellingBee \textit{probe} (using a different random initialization) to the learned embeddings, which yields 93.5\% EM on held-out token types.}
We train RoBERTa-Large on English Wikipedia using the hyperparameter configuration of 24hBERT \cite{izsak2021train}, and cease training after 24 hours (approximately 16,000 steps).
For comparison, we train exactly the same model with a randomly-initialized embedding layer.

Figure~\ref{fig:pretraining} shows the masked language modeling loss with and without pretrained embeddings.
We see that the curves quickly converge into one.
After only 1000 training steps, the difference between the validation losses never exceeds 0.01.
This result indicates that in this scenario, the model does not utilize the character information injected into the tokens' embeddings.

Although there are many possible ways to explicitly add orthographic information to tokens embeddings, our method is relatively straightforward as it gives the model a chance to utilize pre-stored character information. Along with the results from Section \ref{sec:results}, we hypothesize that the implicit notion of spelling that the model learns during pretraining might be sufficient for masked language modeling.


\section{Conclusion}

This work reveals that pretrained language models learn, to some extent, the character composition of subword tokens.
We show that our SpellingBee probe can spell many vocabulary items using their uncontextualized embedding-layer representations alone.
Trying to explicitly infuse character information into the model appears to have a minimal effect on the model's ability to optimize its language modeling objective, suggesting that the model can independently learn all the character-level information it needs for the task.

\section*{Acknowledgements}
This work was supported by the Tel Aviv University Data Science Center, Len Blavatnik and the Blavatnik Family foundation, the Alon Scholarship, Intel Corporation, and the Yandex Initiative for Machine Learning. We thank Avia Efrat for his valuable feedback.

\bibliographystyle{acl_natbib}
\bibliography{anthology,acl2021}

\newpage
\appendix
\section{Levenshtein Distance}
\label{sec:levenshtein}

Levenshtein distance \cite{levenshtein1966binary} is an edit distance metric that, given two strings, calculates the minimal number of changes needed to be done in order to make the two strings identical.
Levenshtein distance \textit{ratio} is the length-normalized version, which is computed by adding the sum of lengths of both strings to the edit distance and dividing by the same sum of lengths. 
We report the main experiment's results using this ratio in Table~\ref{tab:levenshtein}.

\begin{table}[ht]
\centering
\small
\begin{tabular}{@{}lcccc@{}}
\toprule
\multirow{2}{*}{\textbf{Filter}} & \multicolumn{2}{c}{\textbf{RoBERTa}} & \textbf{GPT2} & \textbf{AraBERT} \\
& \textbf{Base} & \textbf{Large} & \textbf{Medium} & \textbf{Large} \\
\midrule
None & 69.7 & 72.7 & 74.4 & 83.6 \\
Similarity & 61.5 & 63.7 & 64.5 & 75.8 \\
Lemma & 61.4 & 63.3 & 63.7 & 74.8  \\
\midrule
Control & ~~25.6 & ~~26.4 & ~~27.0 & ~~25.7  \\
\bottomrule
\end{tabular}
\caption{Levenshtein distance ratio. The first three rows reflect different methods for filtering the training data, and the fourth represents the control experiment, which uses randomly initialized embeddings. All SpellingBee instances in this table are trained on 32000 examples.}
\label{tab:levenshtein}
\end{table}

\section{Spelling Accuracy by Frequency}
\label{sec:token_freq}

\begin{figure}[ht]
\centering
\includegraphics[width=\columnwidth]{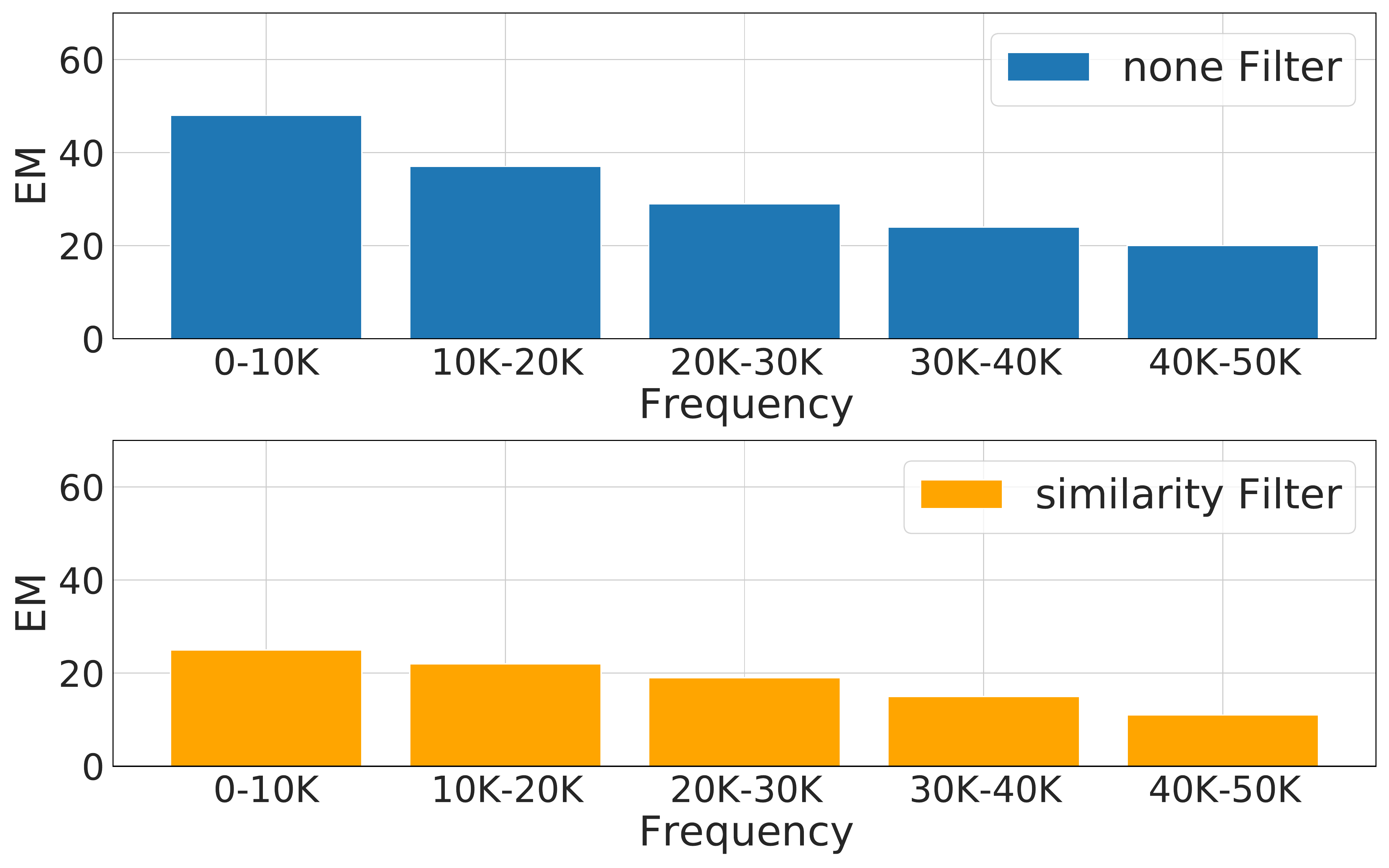}
\caption{The EM scores of SpellingBee on RoBERTa-Large for each frequency quintile with the \textit{none} filter (top) and the \textit{similarity} filter (bottom).
}
\label{fig:freq_analysis_EM}
\end{figure}

\begin{figure}[ht]
\centering
\includegraphics[width=\columnwidth]{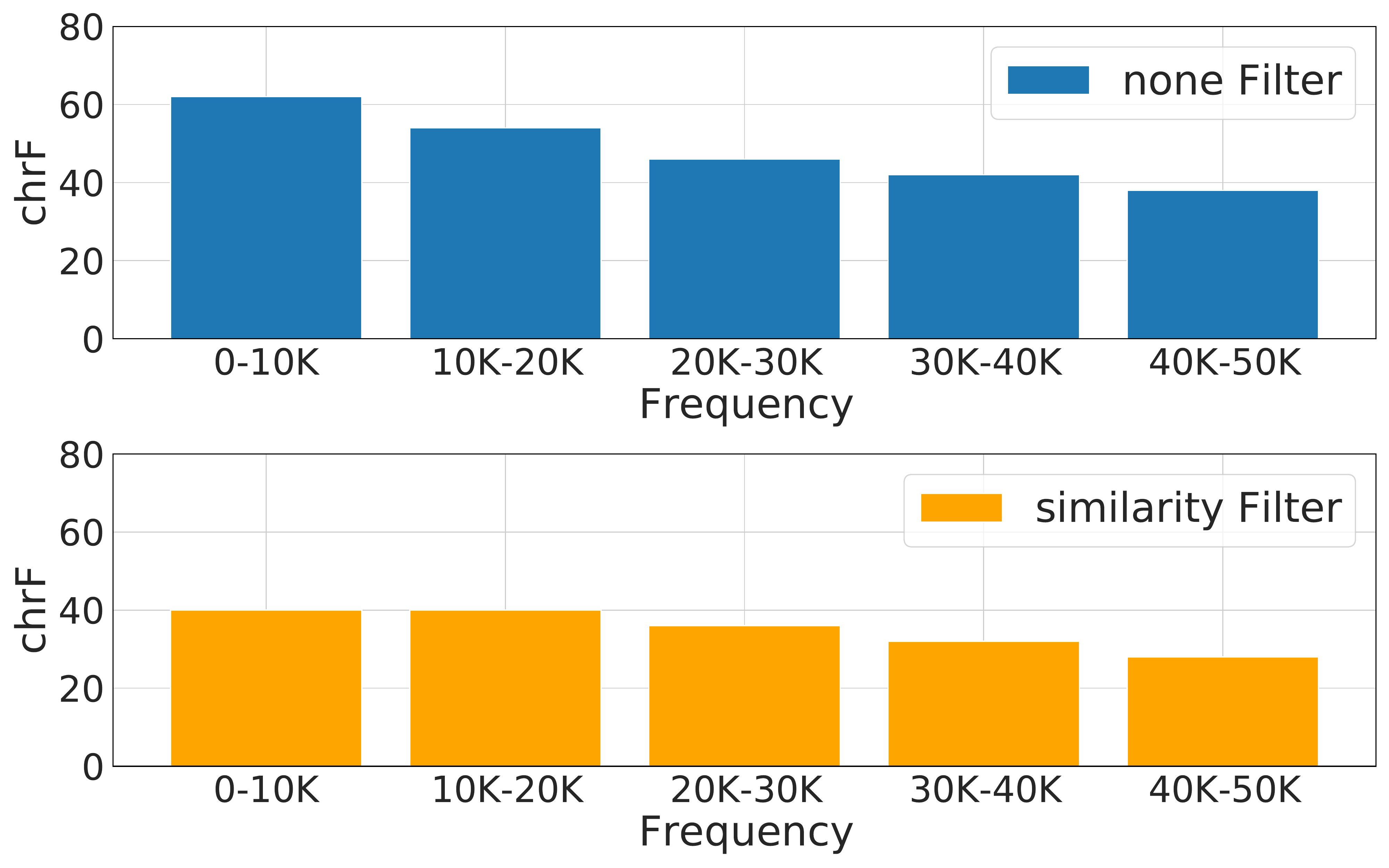}
\caption{The chrF scores of SpellingBee on RoBERTa-Large for each frequency quintile with the \textit{none} filter (top) and the \textit{similarity} filter (bottom).
}
\label{fig:freq_analysis_chrF}
\end{figure}

We test whether pretrained models tend to store more spelling-related information in higher-frequency token types.
We focus on RoBERTa-Large, and assign each token in the test set to its frequency quintile according to the number of times it appeared in the pretraining corpus -- from the 10000 most frequent token types (top 20\%) to those ranked 40000-50000 in the vocabulary (bottom 20\%) -- and measure the average performance of SpellingBee within each quintile.
Figures \ref{fig:freq_analysis_EM} and \ref{fig:freq_analysis_chrF} shows the results with and without the similarity filter.
We observe that SpellingBee is indeed able to extract more information from higher-frequency token types, suggesting that the pretrained model has more information about their character composition.

\section{Spelling Accuracy by Length}
\label{sec:token_len}

\begin{figure}[ht]
\centering
\includegraphics[width=\columnwidth]{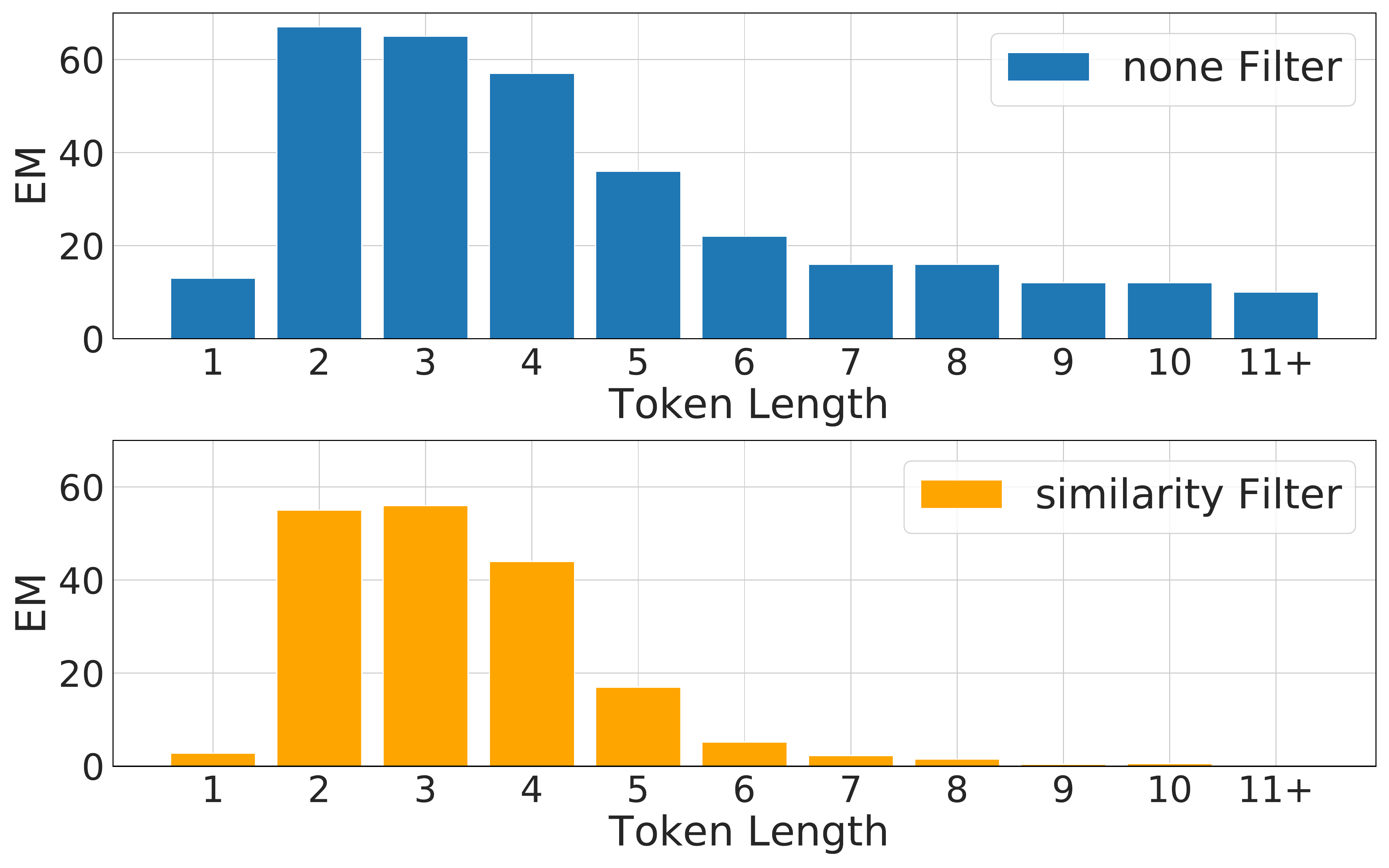}
\caption{The EM scores of SpellingBee on RoBERTa-Large for each token length with the \textit{none} filter (top) and the \textit{similarity} filter (bottom). The rightmost column groups together tokens with length of 11 or above.}
\label{fig:len_analysis_EM}
\end{figure}

\begin{figure}[ht]
\centering
\includegraphics[width=\columnwidth]{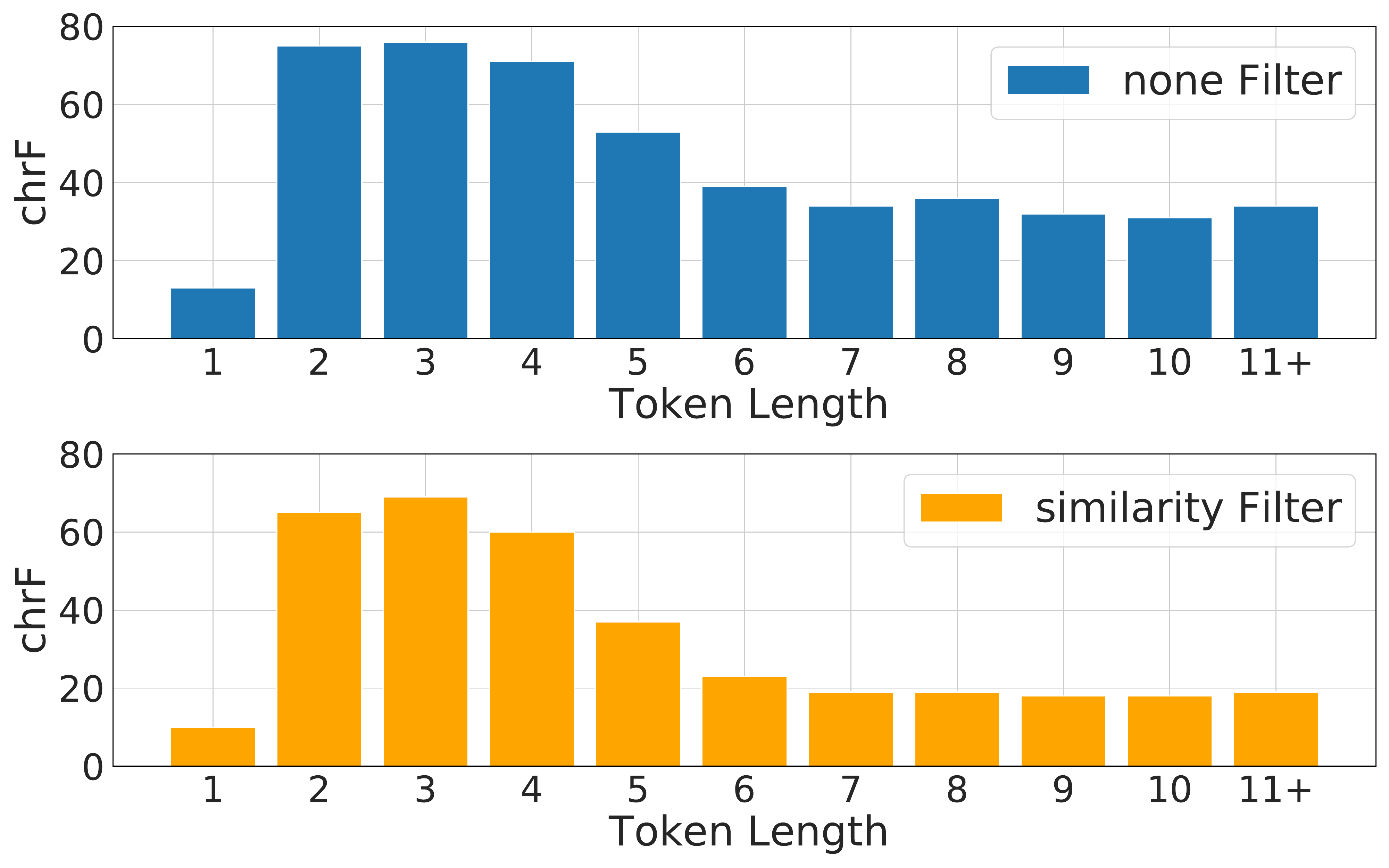}
\hfill
\caption{The chrF scores of SpellingBee on RoBERTa-Large for each token length with the \textit{none} filter (top) and the \textit{similarity} filter (bottom). The rightmost column groups together tokens with length of 11 or above.}
\label{fig:len_analysis_chrF}
\end{figure}

We analyze the effect of token length on the probe's ability to spell.
A priori, it is reasonable to assume that it is easier for the probe to spell shorter tokens, since less information needs to be extracted from the embedding and there are less discrete decisions to be made while decoding.
Indeed, Figure \ref{fig:len_analysis_EM} shows that with the none filter most vocabulary tokens with 2-4 characters can be accurately reproduced from their vector representations, while longer tokens are harder to replicate.
This trend is particularly sharp when the similarity filter is applied, as the probe is hardly able to spell tokens with 6 or more characters accurately; having said that, the probe is able to generate many \textit{partially correct} spellings, as measured by chrF (Figure \ref{fig:len_analysis_chrF}).
Perhaps a less intuitive result is the probe's failure to spell single-character tokens.
A closer look reveals that many of these examples are rare or non-alphanumeric characters (e.g. \textit{ç} and \textit{\$}), which are probably very difficult for the probe to generate if it had not seen them during training.
While these results show strong trends with respect to length, token length is also highly correlated with frequency, and it is not necessarily clear which of the two factors has a stronger impact on the amount and resolution of character-level information stored in the embedding layer of pretrained models.

\section{Manual Error Analysis}
\label{sec:manual_analysis}

We manually analyze 100 random tokens that SpellingBee spelled incorrectly with the lemma filter to understand the nature of the spelling mistakes.
Out of those 100 we display 20 mistakes in Table \ref{tab:errors} alongside the spelling prediction of the control baseline.
SpellingBee's mistakes vary from single-character typos to completely different words.
Having said that, the vast majority of mistakes have significant overlap with the correct spelling, such as shared prefixes and capitalization.

\begin{table}[ht]
\centering
\small
\begin{tabular}{llllll} 
\toprule
\textbf{Token} & \textbf{SpellingBee} & \textbf{Control} \\
\midrule
\_Issa & \_Asey & \_kinston \\ 
\_Rhod & \_Rob & \_hoedn \\ 
Memory & Mathinge & \_entically \\ 
\_metals & \_metrys & \_leaved \\ 
\_Reed & \_Redd  & \_fomparing \\
\_break & \_breach & \_promoters \\
\_summit & \_mosump & \_seasons \\
Catholic & Cravital & \_tonversal \\ 
\_cleanup & \_lamed & \_paclus  \\
\_Winner & \_Womer & \_purden \\
\_LIM & \_LUM & \_Send  \\
Copy & Cople & \_providers \\ 
\_voicing & \_relicing & \_walking  \\
\_Stab & \_Stamb & \_hoviders  \\
\_356 & \_353 & \_budiance  \\
find & wive & \_malding \\ 
\_Psychic & \_Syptanc & \_joacter \\
\_Looking & \_Lowing & parging \\
CLOSE & DEFIC & \_tuldence \\ 
\_prolific & \_promistic & \_complexement  \\
\bottomrule
\end{tabular}
\caption{Sampled SpellingBee errors with the lemma filter alongside the control baseline's spelling for the same tokens. The underscore (\_) represents a preceding whitespace.
}
\label{tab:errors}
\end{table}

\section{Hyperparameters}
\label{sec:hyperparameters}
We implement SpellingBee with a 6-layer encoder-decoder model, with 512 model dimensions.
The model parameters are optimized with Adam \cite{Kingma2015AdamAM} for 1000 steps with up to 1024 tokens per batch, a learning rate of 5e-4, and a dropout rate of 0.1.
These are the default hyperparameters for training a transformer language model in Fairseq \cite{ott-etal-2019-fairseq}.

\end{document}